\documentclass[11pt,a4paper]{article}
\usepackage{times,latexsym}
\usepackage{url}
\usepackage[T1]{fontenc}
\usepackage{booktabs}
\usepackage{graphics}
\usepackage{graphicx}
\usepackage{hyperref}
\newcommand{\zh}[1]{\begin{CJK}{UTF8}{gbsn}#1\end{CJK}}
\usepackage{multirow}
\usepackage{CJK}
\usepackage[acceptedWithA]{tacl2021v1}

\usepackage{xspace,mfirstuc,tabulary}

\newif\iftaclinstructions
\taclinstructionsfalse %
\iftaclinstructions
\renewcommand{\confidential}{}
\renewcommand{\anonsubtext}{(No author info supplied here, for consistency with
TACL-submission anonymization requirements)}
\newcommand{\instr}
\fi

\iftaclpubformat %

\else

\fi

\title{
Finding Challenging Metaphors that Confuse Pretrained \\ Language Models
}

\author{
    Yucheng Li$^\diamond$ 
  \and
  Frank Guerin$^\diamond$ 
  \and
  Chenghua Lin$^\dagger$
  \ \\
  $^\diamond$University of Surrey, UK
  \\
  $^\dagger$University of Manchester, UK
\\
  \texttt{\{yucheng.li, f.guerin\}.surrey.ac.uk}
  \\
  \texttt{chenghua.lin@manchester.ac.uk}
}

\date{}

\begin{document}
\maketitle
\begin{abstract}
Metaphors are considered to pose challenges for a wide spectrum of NLP tasks. This gives rise to the area of computational metaphor processing.
However, it remains unclear what types of metaphors challenge current state-of-the-art models.
In this paper, we test various NLP models on the VUA metaphor dataset and quantify to what extent metaphors affect models' performance on various downstream tasks.
Analysis reveals that VUA includes a large number of  metaphors that pose little difficulty to downstream tasks. We would like to shift the attention of researchers away from these metaphors to instead focus on challenging metaphors.
To identify  hard metaphors, we propose an automatic pipeline that  identifies  metaphors that challenge a particular model. 
Our analysis  demonstrates that our detected hard metaphors contrast significantly with VUA and reduce the accuracy of machine translation by 16\%, QA performance by 4\%, NLI by 7\%, and metaphor identification recall by over 14\% for various popular NLP systems.
\end{abstract}

\section{Introduction}

Metaphors are figurative devices prevalent in language that convey meaning through analogy and comparison \cite{moh-16,rai2020-metaphor-survey}. Empirical studies reveal that around 10\%-20\% of the words in regular parlance are metaphorical \cite{steen2010-from-mip-to-mipvu-most-frequent-verbal-vua,shutova2010-20percents}. 
The non-literal nature of metaphorical language presents challenges for NLP systems as there is an ambiguity between the literal and metaphorical meaning, and the metaphorical meaning might require an advanced level of understanding to truly comprehend. This difficulty has motivated research to handle the identification \cite{krishnakumaran2007-identification,mao2019-mip-spv,choi2021melbert,li2023framebert} and interpretation \cite{veale2008-interpret,su2016-meta-interpretation,mao2018-identification-and-interpretation} of metaphors, leading to increased interest in the area of computational metaphor processing.
Previous works have shown that metaphor processing could benefit a range of real-world NLP applications, such as sentiment analysis \cite{li2023secret}, machine translation \cite{mao2018-identification-and-interpretation}, information retrieval \cite{veale2011-metaphor-retrieval}, and natural language generation systems \cite{zhou2020-love,chakrabarty2020generating,li-etal-2022-nominal,li2022cm}.

\begin{figure}
    \centering
    \includegraphics[width=\columnwidth]{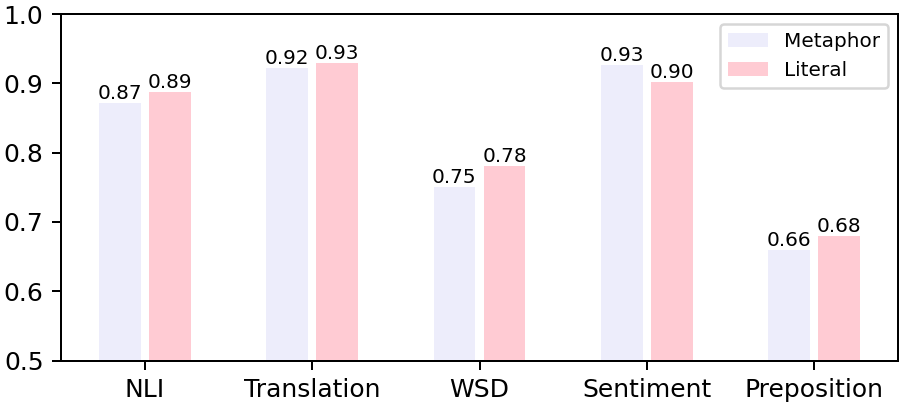}
    \caption{
    Comparison between metaphoric and literal samples from VUA dataset on various NLP tasks.
    }
    \label{fig:glue}
\end{figure}

However, it remains unclear precisely which types of metaphors pose difficulties for modern NLP models. Without this knowledge, metaphor processing research risks focusing on the easier %
metaphors that current NLP systems can already handle, rather than tackling unsolved challenges. To verify this possibility, we  test modern NLP models on metaphor examples from the VUA metaphor corpus across various downstream tasks. We find the majority of VUA metaphors have little impact on downstream performance, containing many simple or conventional metaphors which current models handle effectively. This result is illustrated in Fig~\ref{fig:glue}, with full details in Section 3.

Within  computational metaphor processing research, significant effort has been spent metaphor identification  on the VUA dataset \cite[overviewed in][]{neidlein-etal-2020-analysis,tong2021recent}. However from a downstream task perspective it might not matter whether a word is a metaphor or not, so long as the downstream task is completed correctly. What matters for the downstream task is to identify the correct sense of a word, when metaphorical or literal options are both available. 
This leads us to the idea of finding hard metaphor samples where their sense is easily confused by language models.
These  metaphors would cause degradation in downstream tasks, and focusing research on processing them correctly would lead to valuable advances in metaphor processing techniques.
To this end we propose an unsupervised method to automatically identify hard metaphors for a specific model.
The core of the approach is metaphor word sense disambiguation: determining whether a model can distinguish the metaphorical sense of a word correctly from other senses.
Specifically, we leverage contrastive learning and clustering on an the embeddings from a particular model, to identify metaphors whose true meaning is confused with another based on similarity. These indistinguishable metaphors are considered hard metaphors for the model.

Our approach provides model-tailored insights into metaphor phenomena that continue to challenge NLP systems. We then test model performance on various downstream tasks and show that performance is significantly degraded on sentences with hard metaphors, compared to literal language. Most computational metaphor processing works make use of some existing LLM as a base or foundation, and add extra layers and training schedules to deal with metaphors \citep{mao2018-identification-and-interpretation,mao2019-mip-spv,li-etal-2023-metaphor,wang2023metaphor,li2023metaphor}. Our recommendation is that researchers would first use our method to find the hard metaphors for their base model, and use this as a challenge test set. In this way they can focus on computational metaphor processing which will significantly improve the performance of downstream tasks.

To summarise, our contributions   are as follows: (1) analysis on the VUA metaphor dataset revealing that VUA is mostly populated by  metaphors that pose no problems for modern NLP systems; (2) a novel framework for automatically identifying model-specific hard metaphors based on word sense disambiguation through contrastive learning representations and clustering. %

\section{Metaphors in Linguistics versus NLP}
In this section, we examine the annotation protocols of popular metaphor corpora, analyse the composition of VUA, the most popular metaphor dataset in NLP, and review its impact on NLP tasks.

\subsection{Metaphors Annotation by Linguists}
\label{mip}
Metaphor annotation studies are usually corpus-based, with the identification of metaphorical usages of words or a search for specific metaphors in a corpus.
Early metaphor works focused on limited domains, e.g., metaphors in \textsc{war, economy}, or \textsc{food} \cite{koller2004-meta-in-limited-domains-3,sznajder2005-meta-in-limited-domains-2,hardie2007-meta-in-limited-domains-1}, or specific genres \cite{charteris2000-metaphor-in-economic,klebanov2013-meta-in-test-taker-essays}.
Pragglejaz \citet{group2007-mip} took a more general approach to how metaphor behaves in unrestricted corpora, proposing a Metaphor Identification Procedure (MIP).
In MIP, every word is tagged as literal or metaphorical usage via a three step protocol:\label{bm}
1) Establish the contextual meaning of each word; 2) Determine if the word has a more basic meaning; 3) If the contextual meaning contrasts with the basic meaning, mark the word as metaphorical; if not, mark it as literal. The basic meaning in MIP is not necessarily the most frequent meaning, rather, it is the more ``concrete, body-related, and historically older'' meaning based on dictionaries. \citet{steen2010-from-mip-to-mipvu-most-frequent-verbal-vua} expands and refines  MIP  and %
covers more types of metaphors, e.g., \textit{borderline metaphors, metaphor signals, and personification}.

MIP is used in the most popular metaphor corpus VUA. %
However, some of the words labelled as metaphor in VUA may be surprising to outsiders; e.g., a preposition such as `in' can be labelled metaphoric if it is not the basic meaning of physically moving into a container or bounded area, e.g., `in a political party`, or `in love'. 
Even pronouns, e.g., 
\textit{it, this, these}, can be labelled metaphoric when they refer to a metaphoric word. %

\label{vua_analysis}
\begin{figure*}
    \centering
    \includegraphics[width=\textwidth]{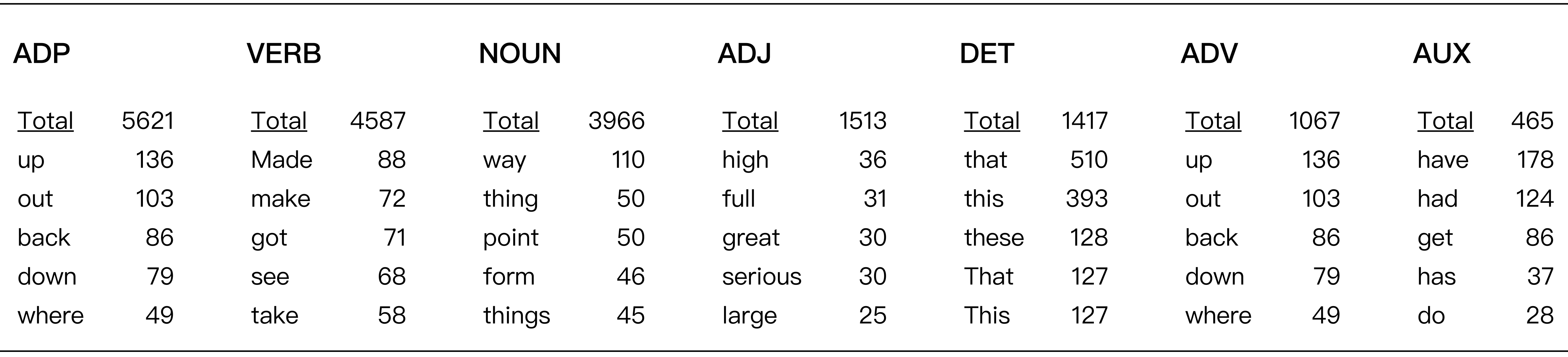}
    \caption{POS breakdown analysis on metaphors from VUA test set.}
    \label{fig:vua-pos}
\end{figure*}

We present a part-of-speech (POS) breakdown analysis on the VUA metaphor dataset to illustrate its composition. We take the VUA test set that contains over 22k metaphorical and literal sentences, which is randomly split from the training set. Each target word was annotated with a universal POS tag. We group the metaphorical sentences based on the universal POS tag of the target metaphorical word and provide statistics and most frequent examples for each POS in Fig. \ref{fig:vua-pos}. Surprisingly, this shows that prepositions (ADP) account for the largest portion of metaphorical words, making up over 25\% of all annotated metaphors. This is followed by verbs (VERB), nouns (NOUN), and determiners (DET). The prevalence of prepositions and determiners is unexpected, since people often do not regard  them as metaphorical.
Looking at frequent examples within each POS category, we discover many conventional or suspicious metaphors. 
For instance, common prepositional metaphors include "in debt", "on track", "under control"; frequent metaphorical determiners like "that point", "this way", "those things";  the most common metaphorical verbs and nouns also appear highly conventional. 
This is because MIP includes a word as a metaphor as long as it is not using its very basic and historical meaning, which will technically include a very broad range of metaphors.
In the remainder of this paper we will use `MIP metaphor' to refer to this broad notion of metaphor.

Another way to identify metaphor is to distinguish metaphorical word senses, which we call Metaphorical Sense Identification Procedure. \citet{kilgarriff1997-rethink-word-sense} finds metaphor to be a ubiquitous source of polysemy, and a systematic relation exists between metaphor and word sense.
This is not to say that all metaphors are already condensed as word senses: many creative metaphors are not recognised word senses. However truly novel metaphors are extremely rare in everyday human language \cite{steen2010-from-mip-to-mipvu-most-frequent-verbal-vua} (more common in poetry), and  the vast majority of metaphors encountered are recognised word senses.
\citet{lonneker2004-meta-sense} investigate metaphor annotation in lexical resources via Metaphorical Sense Identification Procedure. Their Hamburg Metaphor Database maps metaphorical expressions to senses from EuroWordNet. \citet{moh-16} present a study to compare the emotionality of metaphorical expressions and their literal counterparts, where metaphorical uses are annotated using by Metaphorical Sense Identification Procedure.
The broader metaphor notion in MIP, than \citet{moh-16},  shows the disagreement between linguists and the NLP community about what types of metaphor are of interest.
We side-step the linguistic debates about what is or is not a metaphor, by  focusing only on metaphors that cause practical problems for NLP downstream tasks.
This is also close to the point  \citet{kilgarriff1997-rethink-word-sense} made about word senses: that they exist only relative to a task.

\subsection{MIP Metaphors in NLP}

Computational metaphor processing  aims to pre-process metaphors to improve NLP performance, based on the hypothesis that: general NLP techniques are more likely to fail in dealing with metaphorical expressions. This motivates the  metaphor identification and interpretation tasks, however,  metaphor identification mostly tackles MIP metaphors, where  we have seen (Fig.~\ref{fig:glue}) that the majority of them do not necessarily cause problems for NLP applications (see \S\ref{glue} for more detail). For example, in the metaphor \textit{``\textbf{valuable} experience''}, the \textbf{valuable} here is labelled as metaphor (by MIP) since its contextual meaning \textit{``useful and important''} is not the most basic sense \textit{``worth a lot of money''}. However, based on the statistics from SemCor \cite{miller1994-semcor}, a WSD dataset, \textit{`very useful and important'} is the most frequent sense of \textbf{valuable}, and therefore it might not trouble statistics-based NLP methods. The same also applies to these metaphors identified via Metaphorical Sense Identification Procedure, e.g., \textit{`\textbf{sweet} boy'}, or \textit{`\textbf{way}'} meaning method, or \textit{`\textbf{back}'} meaning a direction, and not the basic meaning `rear surface of the human body'.
The phrase `\textit{stumbled upon}' is so frequently used metaphorically that machine translators tend to render it as `found' or `discovered'. A literal use such as 
`While walking through the forest she stumbled upon the root of a tree, and fell.' is translated into Chinese as `\ldots she discovered the root of a tree, and fell.'\footnote{Example %
tested %
in Google Translate on 24.08.2022.} So we see that metaphors that are very common in a training set may not need any metaphor processing; instead,  their more rare literal counterpart  might need some processing help.

We propose that metaphor processing should tackle challenging metaphors which current language processing techniques struggle with, and which lead to errors in real applications. There are two possible solutions for discovering these `hard metaphors'. \textbf{First}, we could find metaphors that are hard for each downstream task. In translation, \textit{``launder money''} is not a hard metaphor, because literally translating the metaphorical word \textit{``launder''} to many target languages can still obtain a valid metaphor, such as \textit{``blanchir d'argent''} in French and ``\zh{洗钱}'' in Chinese. In the metaphor \textit{`This rare Bordeaux must be allowed to \textbf{breathe} for at least 2 hours.'}, the literal translation of \textbf{breathe} results in failure in a Chinese translation, and while it results in an acceptable French translation,  `a\'erer' is more common. Thus, we can speculate that metaphors whose literal translations do not form a valid metaphor in a target language are hard metaphors for machine translation systems. But different criteria might apply for other downstream tasks.
\textbf{Second}, 
considering the statistical nature of popular language processing methods, creative or rare metaphorical expressions might be  difficult for NLP . For instance, %
a common word sense of \textit{`hot'}  such as \textit{`a hot topic', `a hot argument'} is less likely to cause problems, but creative expressions which occur rarely in language corpora, e.g., \textit{`phones rang hot', 'nights over a hot typewriter'} are harder for a machine to understand.

We will focus on the second solution: to find hard metaphors by doing a case level analysis (where a case is the occurrence of a word in context), to find cases where SOTA methods fail. We leave the first solution for future work.

\begin{table*}[t]
    \centering
    \resizebox{\textwidth}{!}{
    \begin{tabular}{l|l}
    \toprule
        Task & Examples \\
        \midrule
         NLI & S1: This is also the \textit{\underline{view}} put forward by John Harris. \\
         & S2: This is also the \textit{\underline{opinion}} put forward by John Harris. \\
         & \textbf{Answer}: Equivalent \\
         \midrule
         Translation & Source: their claim may only amount to \textit{\underline{around}} 11,000 pounds. \\
         & Target: \zh{他们的索赔可能只有11,000英镑\underline{左右}} \\
         \midrule
         WSD & The struggle \textit{\underline{reflects}} a shift in public awareness about the need for action to change reality. \\
         & \textbf{Choices}: A) to throw or bend back (from a surface); B) manifest or bring back; \\ 
         & C) reflect deeply on a subject; D) give evidence of a certain behaviour; .. \\
         & \textbf{Answer:} D \\
         \midrule
         Sentiment & You are becoming so \textit{\underline{hard}} and bitter and it 's not really like you . \\
         & \textbf{Answer}: Negative \\
         \midrule
         Preposition & Urban Development Corporations were introduced \textit{\underline{in}} 1981. \\
         Comprehension & \textbf{Choices:} A) Location; B) Time; C) Direction; D) Manner; E) Cause or reason; \\
         & F) Agent; G) Possession; H) Comparison \\
         & \textbf{Answer:} B \\
         \bottomrule
    \end{tabular}}
    \caption{Examples of the NLP tasks created with VUA metaphors.}
    \label{tab:task_examples}
\end{table*}

\section{Impact of VUA Metaphors on Various NLP Tasks}\label{glue}

Automated metaphor processing is important for a number of NLP applications, such as document summarisation, question answering, and text style transfer \cite{tong2021recent,li-etal-2020-dgst}. However, the extent to which  metaphors upset downstream tasks, and which types of metaphors are most challenging, remain unclear. Here we construct five NLP tasks for VUA metaphors and quantitatively analyse the impact of the metaphors on the tasks. 

\noindent\textbf{Natural Language Inference (NLI).} In NLI tasks, the goal is to determine whether a pair of sentences convey the same meaning or not. As shown in Table \ref{tab:task_examples}, the pair consists of a VUA metaphor as S1, and annotators manually paraphrase the metaphorical sentence to a literal sentence as S2. This ensures the metaphor itself is the key to answer the NLI question correctly.

\noindent\textbf{English to Chinese Translation.} We use a metaphorical sentence from VUA as the source, and use translation systems to generate the target translation. Manual evaluation is followed to verify whether the metaphor is correctly translated.

\noindent\textbf{Word Sense Disambiguation (WSD).} This task is to identify which is the correct word sense the VUA metaphor uses. For the example in Table \ref{tab:task_examples}, \textit{reflect} has many word senses in WordNet, we take these as the choices and the task is to find the correct sense of \textit{reflect}, given its context.

\noindent\textbf{Sentiment Analysis.} Metaphor is believed to have a close connection with sentiment. This task is to predict whether a metaphor conveys a positive or negative sentiment. We focus on adjective and adverb metaphors here, as they are more likely to determine the sentiment conveyed by a sentence.

\noindent\textbf{Preposition Comprehension.} %
Normally, ADP words are considered as \textit{stop words} as they are least important in language understanding. A standard solution for ADP words is deleting them before downstream processing. However, as ADP words constitute a significant portion of VUA metaphors, we design this task to test whether NLP systems can understand ADP metaphors correctly. This task aims to identify  the true relation indicated by a preposition. As in the Preposition example in Table \ref{tab:task_examples}, the literal meaning of \textit{in} indicates the location relationship, so \textit{in 1981} is considered a metaphor in VUA. %
We involve eight types of relationships, which cover most of the cases for prepositions.

To assess the impact of metaphor on these tasks, we need an equal number of literal samples to compare with. We construct corresponding literal tests using literal sentences from the VUA dataset.
These literal tests are designed to parallel the metaphorical tests in terms of word choice but diverge in the context where these words are deployed.
For instance, if the metaphorical test focuses on the word `view' in a metaphorical sense, as illustrated in the NLI task in Table \ref{tab:task_examples}, then the corresponding literal test will also feature the word `view', but in a context where it is used literally. Here is the literal sample:

\noindent S1: And then we \textit{\underline{view}} the sun rise (from VUA)

\noindent S2: And then we \textit{\underline{see}} the sun rise (constructed)

\subsection{Experiments Setting}

\noindent\textbf{Models.} We employ several popular NLP models on the five tasks. First, we utilise fine-tuned RoBERTa \cite{liu2019roberta} models on our NLI and sentiment test. The RoBERTa models are fine-tuned on a popular NLI dataset SNLI and sentiment analysis dataset SST-2. For the English to Chinese translation task, we use Google Translate. We use \texttt{wsd-biencoder} for the WSD task. As there is no existing solution for preposition comprehension, we use \texttt{gpt-3} as a zero-shot approach to identify the correct relation.

\noindent\textbf{Data construction and annotation.} For the first four tasks we randomly selected 200 metaphor and 200 literal samples from VUA. For the final Preposition task we used 100 samples of each. For very long samples in VUA we do not take the whole context, but only the sentence with the target word. The translation, sentiment analysis, and preposition comprehension tasks require no further efforts for the question construction. For the other  tasks, we need to create the questions manually. For NLI tests, the S1 is sampled from VUA, but the S2 is manually paraphrased literal sentence of S1. For the WSD tests, we retrieve all word senses of the target word from WordNet as the candidate choices. Five crowd sourcing workers are employed for the annotations. They were presented with general annotation guidelines with question examples, and each question and answer are annotated by at least three workers. We calculated the inter-annotator agreement score, using Krippendorff's Alpha, for labels in our downstream annotation and obtained a value of 0.67, indicating good agreement.

\begin{table}[]
    \centering
    \resizebox{\columnwidth}{!}{
    \begin{tabular}{lcccccc}
    \toprule
         & Num. & NLI & Translation & WSD & Sentiment & Preposition \\
    \midrule
        Metaphor & 900 & 0.87 & 0.92 & 0.75 & 0.93 & 0.66\\
        - VERB & 272 & 0.85 & 0.93 & 0.77 & 0.90 & -\\
        - NOUN & 240 & 0.89 & 0.93 & 0.73 & - & -\\
        - ADJ & 152 & 0.81 & 0.90 & 0.65 & 0.94 & -\\
        - ADV & 123 & 0.90 & 0.89 & 0.80 & 0.94 & -\\
        - ADP & 100 & - & - & - & - & 0.66 \\
    \midrule
        Literal & 900 & 0.89 & 0.93 & 0.78 & 0.90 & 0.68\\
    \bottomrule
    \end{tabular}}
    \caption{Comparing accuracy for literal or MIP metaphor samples for  five NLP tasks.}
    \label{tab:vua_results}
\end{table}

\subsection{Results}\label{vuametaphorresult}

Table \ref{tab:vua_results} indicates that the impact of VUA metaphors on various NLP tasks is relatively small. 
The differences between metaphor (average) and literal range from 1\% to 3\% across the five tasks, i.e. very small differences, with the largest being WSD and sentiment analysis where metaphors may cause more confusion.
We expect the differences are due to a small proportion of VUA metaphors being hard. 

When dissected by part-of-speech (POS) tags: VERB, NOUN, ADJ, ADV, and ADP, the performance gaps remained minimal. For example, VERB metaphors scored 0.85 in NLI, closely following the overall metaphor score. This pattern held across other POS categories as well, suggesting that even when considering the type of metaphor involved, the impact on NLP tasks is not significantly different from that of literal language.

Overall, the results indicate that advanced NLP models like RoBERTa and GPT-3 are increasingly capable of handling metaphorical language, albeit with room for specific improvements.

\begin{figure*}
    \centering
    \includegraphics[width=\textwidth]{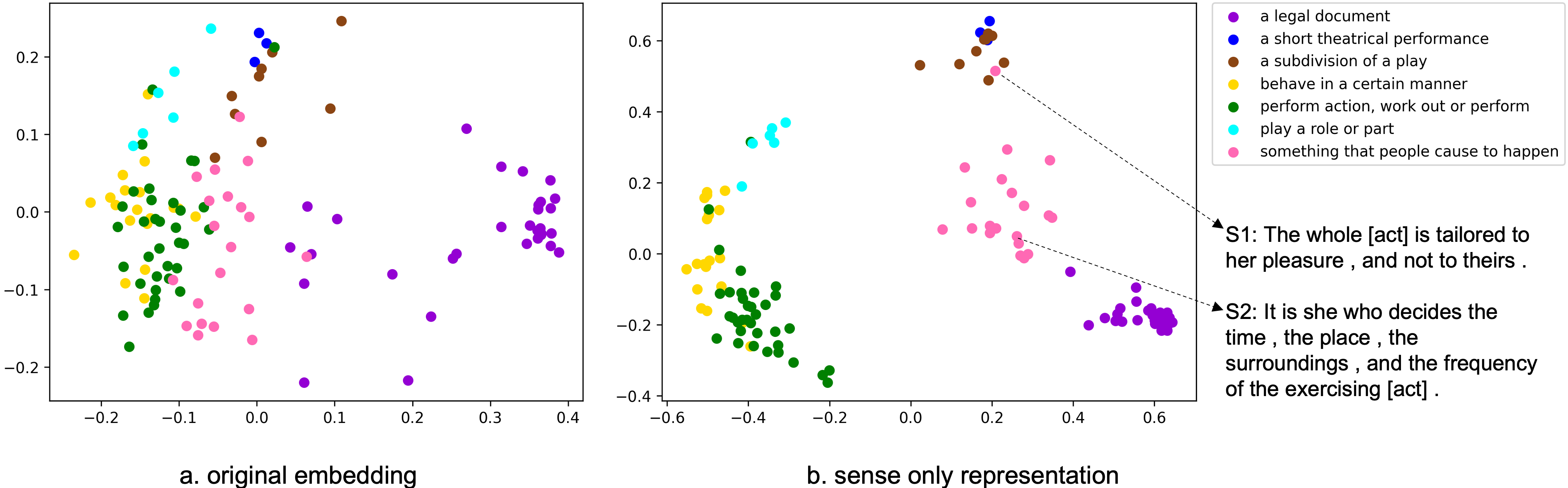}
     \caption{PCA visualization of RoBERTa's original embedding (a) and  Sense Only Representations (SORs) (b) of the word \textit{\underline{act}} in different passages. The legend shows the  word sense gloss from WordNet. Two examples are given at the bottom right,  one (S2) distinguished successfully, the other (S1) in a wrong sense cluster. Examples are from the SemCor dataset. }
    \label{fig:scatter}
\end{figure*}

\section{Identifying Hard Metaphors}

The hard metaphor identification pipeline consists of two stages: 1) obtaining sense-only representation based on contrastive learning; 2) identifying hard metaphor examples from a WSD dataset using a clustering  algorithm.

\subsection{Overall Procedure}
\label{overall}
The task of distinguishing a word sense from others is called word sense disambiguation (WSD), often used as an NLU benchmark \cite{navigli2018-nlu-wsd}. We propose to probe PLMs' understanding of metaphors via a WSD task by testing how well can PLMs' representations disambiguate metaphorical word senses from others.
For example, if the \textit{hot} in \textit{``a hot topic''} is identified as the sense of \textit{high in temperature} rather than \textit{exciting or popular} by a NLU system, then the NLU system misunderstood the  meaning of the metaphor here.

We cannot directly use a fine-tuned WSD model to say that metaphors that failed to be classified are hard metaphors, because even the largest WSD dataset, SemCor, is quite sparse and does not cover many metaphorical word senses at all, while often having few examples for some senses. That means some metaphors might fail to be tackled by a fine-tuned WSD model just because of the bias of the WSD training set.

Therefore, we  formulate the WSD procedure as a clustering problem using vanilla PLMs' representation. According to \citet{wiedemann2019-sense-prob}, contextualized word embeddings in different contexts tend to gather together based on their word senses in the vector space. Thus hard metaphors could be detected by finding metaphoric expressions that fail to be clustered with cases sharing the same word sense. Moreover, since PLMs are trained on huge corpora, their word representations are more close to the real word distributions of language, which makes the hard metaphor identification robust.

Specifically, we assume that a NLU system fails to distinguish an example $(g, w, S)$, which consists of word sense $g$ of word $w$ in passage $S$, if its contextual embedding $v$ is placed close to the cluster of other word senses of $w$ in the vector space. In the  example shown in the right  of Fig.~\ref{fig:scatter},  RoBERTa misplaced the \textit{\underline{act}} within S1 to a wrong cluster of the word sense \textit{``a subdivision of a play''} rather than the reference sense \textit{``something people cause to happen''}, which shows that RoBERTa confused the lexical semantics of S1. A correctly tackled example should be like S2, which is placed near examples sharing the same meaning.

However, the original embedding produced by PLMs should not be used directly,
as the representations of a word with the same word sense differ, due to different word positions and context information \citep{rogers2020-bert-prob}.
To reduce the influence of position and context information, while preserving  the word sense information, we employ a contrastive learning \cite{chen2020-simclr} approach that brings the contextual word embeddings with the same word sense as close as possible together and pushes embeddings of different word senses  far away. We call the obtained word representations Sense Only Representations (SORs). As shown in Fig.~\ref{fig:scatter}, SORs form clearly separable clusters, compared to the original embedding.
Our later results in \S\ref{downstream result, varphi} show that more overlapped clusters in the SOR space are strongly correlated with failures of PLMs.

To quantify the distinguishability, we project a list of examples $(g, w, S)$ of word $w$ to vector space. For each example denoted as $(g_i, w_t, S_j)$, we define an \textit{overlap ratio} $\varphi = s/k$, where $s$ is the number of examples sharing the same word sense $g_i$ in the $k$ closest neighbours (details in \S\ref{clustering}). Higher $\varphi$ means the specific example is well distinguished, lower  indicates it is not easily distinguished. We treat examples whose $\varphi$ is lower than a threshold $\xi$ as challenging examples, and hard metaphors if the example is used metaphorically.

In the rest of section, we describe in detail how we obtain SORs in \S\ref{sor}. Then we formulate how we compute the overlap score $\varphi$ in \S\ref{clustering}.

\begin{figure}
    \centering
    \includegraphics[width=0.7\columnwidth]{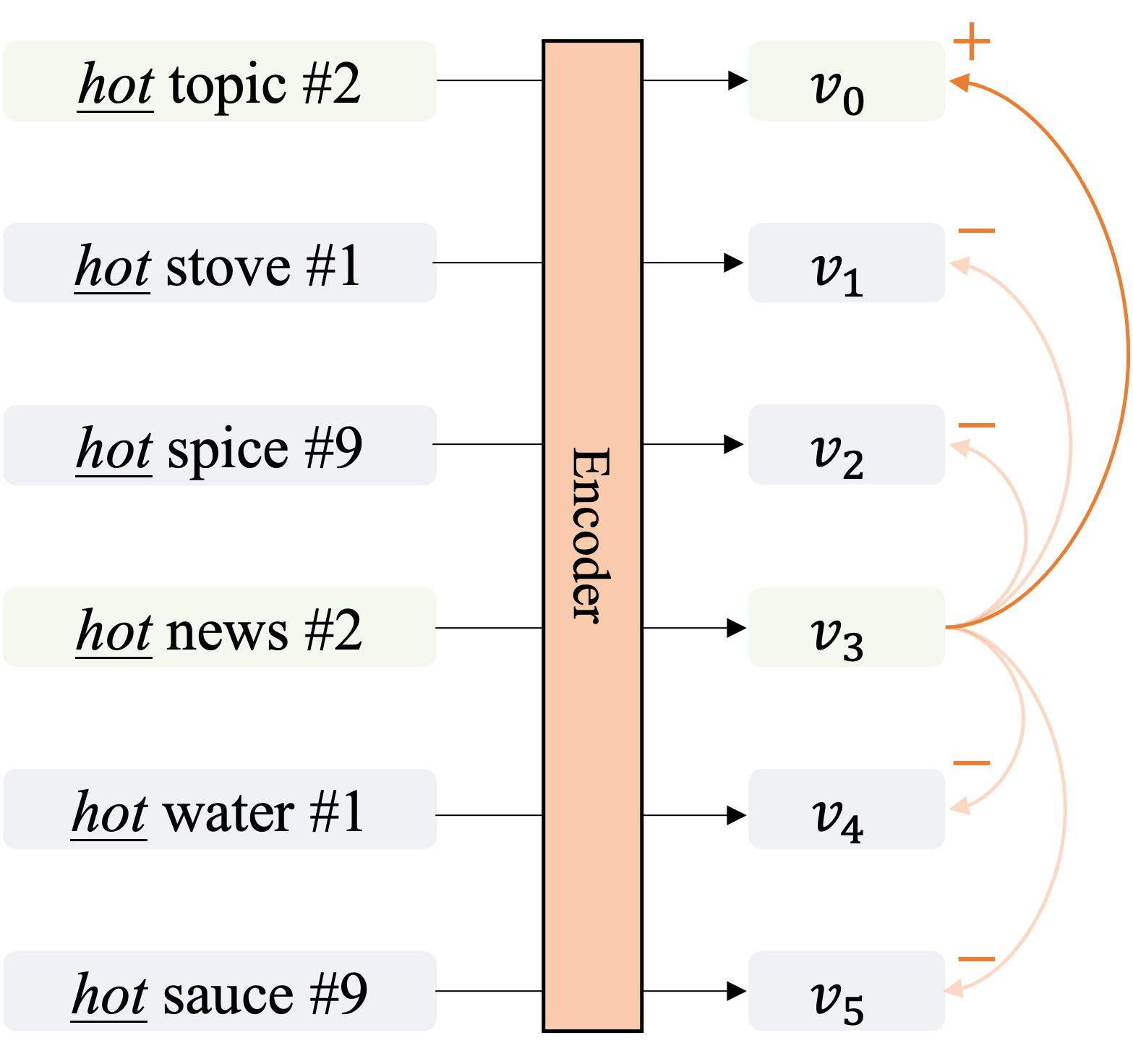}
    \caption{Contrastive learning on word senses.
    }
    \label{fig:cl}
\end{figure}

\subsection{Contrastive Learning on Word Senses}
\label{sor}

As stated above we use contrastive learning to obtain Sense Only Representations (SORs).
We use the contrastive framework in \citet{chen2020-simclr} and take a cross-entropy objective with in-batch negatives \cite{chen2017-negative-sampling,gao2021-simcse}. Specifically, for each instance $x$ of word $w$ and word sense $g$, we sample another example with the same word and word sense to form a positive pair $(x_i, x_i^+)\sim g_i$. Here both instances belong to the word sense $g_i$. To obtain negatives, we sample instances $(x_j, x_j^+)\sim g_j, j\neq i$ with another word sense $g_j$ to extend $(x_i, x_i^+)$ to $(x_i, x_i^+, x_j, x_j^+, ...)_{N}$, where $N$ indicates batch size and is an even number. In each batch, therefore, there will be two instances belonging to the same word sense, and the other $N-2$ instances are treated as negatives for the sense, as in Fig.~\ref{fig:cl}. The training objective $\mathcal{L}$ is defined as follows:
\begin{equation}
    -\log \frac{e^{\mathrm{sim}(v_i, v_i^+)/\tau}}{\sum^{N-2}_{j=1}e^{\mathrm{sim}(v_i, v_j^-)/\tau} + e^{\mathrm{sim}(v_i, v_i^+)/\tau}}
\end{equation}
where $v$ denotes word representations produced by the PLMs, sim() function is realised by cosine distance, $\tau$ denotes a temperature hyperparameter.

\subsection{Computing Overlap Ratio}
\label{clustering}

Following the motivation in \S\ref{overall}, we assume that a metaphorical SOR that is close to a different sense cluster in vector space is a hard metaphor. To measure the distinguishability for each sample $(g, w, S)$, we use a metric named \textit{overlap ratio} $\varphi = s/k$ which computes the ratio of instances sharing the same word sense $g$ in its $k$ nearest neighbours. The $s$ here indicates the number of neighbours sharing the same word sense $g$. In our experiments, $k$ is set to the number of instances with same word sense $g$.

In Fig.~\ref{fig:scatter}, examples S1 and S2 give a good illustration of how overlap ratio measures to what extent each sample is being correctly placed in the vector space. 
For S1,  most of its neighbours do not share the same sense, so overlap ratio is quite low, $\varphi_{s1} = 1/19$. In contrast, S2 is placed in the cluster of its word sense, so $\varphi_{s2}$ is  higher at $\varphi_{s2} = 18/19$.

\section{Construct A Hard Metaphor Dataset}
Here we describe the implementation details of our hard metaphor identification procedure and  how we tailor a hard metaphor dataset (HMD) for RoBERTa, a typical Transformer-based  model.

\begin{table*}[ht]
    \centering
    \resizebox{\textwidth}{!}{
    \begin{tabular}{lp{3cm}llp{12cm}}
    \toprule
        Term & Word sense & Class & $\varphi$ & Passage \\
        \midrule
        flow & move freely as if in a stream & meta & 0.42 & Yet within this limitation there is an astonishing variety : design as intricate as that in the carpet or miniature , with the melodic line like the painted or woven line often \textit{\underline{flowing}} into an arabesque . \\
        & liquids, move along & literal & - & His first shot in the Open last year landed in a brook that \textit{\underline{flowed}} along the right side of the fairway . \\ \midrule
        push & make publicity for; try to sell & meta & 0 & Governments may fail to \textit{\underline{push}} outward to win more and more people to the national effort , becoming instead rigid and inflexible in their policies. \\
        & move with force & literal & - & The cleaner air means less time spent \textit{\underline{pushing}} a vacuum , fewer trips to the dry cleaners , lighter loads for the washing machine. \\ \midrule
        sit & be in session & meta & 0 & Anyone with a wide acquaintance in both groups and who has \textit{\underline{sat}} through the many round tables, workshops or panel discussions. \\ 
        & be in session & meta & 0.6 & Said Karns, who is a City judge in East St. Louis \textit{\underline{sitting}} in Cook County court. \\ \midrule
        reflect & manifest or bring back & meta & 0.55  & She glanced at the man nodding beside her , a man with weather cracks furrowed into his lean cheeks , with powdery pale eyes \textit{\underline{reflecting}} all the droughts he had seen. \\
        & manifest or bring back & meta & 0.78 & An example may be seen in the Southern Negro 's quest for a position in the white-dominated society , a problem that has been \textit{\underline{reflected}} in regional fiction especially since 1865.\\
        \bottomrule
    \end{tabular}}
    \caption{Hard metaphor examples from HMD. Underlined  words  are target words. $\varphi$ is overlap ratio. Lower $\varphi$ means harder to distinguish metaphor from literal.}
    \label{tab:examples}
\end{table*}

\subsection{WSD Data and Metaphor Annotation}

We use WSD data from UFSAC \cite{vial2018ufsac}, a unification of popular WSD corpora including SemCor, SensEval, SemEval and more.
More dataset statistics are shown in Table~\ref{tab:dataset}.

The Hard metaphor identification also needs metaphor labels for sense inventories. Existing metaphor annotations on word sense inventories, such as MOH metaphor dataset, mainly focus on verbs and exclude metaphors with other POS tags. Therefore, we perform our own metaphor annotation on word senses to cover a wider variety of metaphors. We extend the MOH annotation by adding metaphor labels for word sense inventories of nouns, adjectives, and adverbs, following the same annotation protocol. Specifically, we chose words with at least three senses, so that there is a higher chance of at least one sense being metaphorical, and less than ten senses, to avoid highly ambiguous words. Three native English speaking college students were hired for the annotation. They were given annotation guidelines along with several annotated examples and a video demonstration. We call our annotation Metaphorical Word Senses  (MWS) and its statistics are shown in Table \ref{tab:dataset}. There are 714 metaphorical word senses found among 2955 senses in our MWS. The inter annotator agreement score (Krippendorff's Alpha) is 0.51. In the case of disagreement a simple average was taken.

\begin{table}[]
    \centering
    \resizebox{\columnwidth}{!}{
    \begin{tabular}{lcccc}
    \toprule
    & UFSAC & MOH & MWS & HMD \\
    \midrule
        \#word & 56k & 440 & 671 & 82 \\
        \#sense & 93k & 1638& 2955 & 110 \\
        \#example & 2.9M & - & - & 21k \\
        \#sense per word & 4.2 & 3.7 & 4.4 & - \\
        \#example per word & 18.9 & - & - & 263\\
        \#token per example & 35.8 & - & - & 134 \\
        \bottomrule
    \end{tabular}}
    \caption{Statistics of UFSAC, MOH, MWS, and HMD.}
    \label{tab:dataset}
\end{table}

\subsection{Model and Experimental Setting}
In our experiments, hard metaphors are tailored for the RoBERTa model.
RoBERTa is a typical Transformer-based pre-trained language model and maintains a good performance on various NLP tasks, making it a good  representative model for further analysis. We take SemCor \cite{miller1994-semcor}, the largest corpus inside UFSAC, as the training set for contrastive learning. To avoid any bias, we do not involve any metaphoric senses in the contrastive learning procedure as we let the model do inference on metaphoric instances. 
We set the batch size as 64 and max\_length of passage as 128. We observe that the model converges quickly after 900 learning steps in contrastive learning, so we choose the model checkpoint of 900 steps in our further experiments. %
The  overlap ratio $\varphi$ is then calculated for each instance.
Based on empirical analysis we consider instances with overlap ratio lower than 0.8 as hard metaphors in our experiments. %
To ensure the overlap ratio is statistically valid, we only consider word senses having at least four examples occurring in the WSD dataset.

\subsection{Hard Metaphor Dataset Overview}
We find 21k hard metaphor examples in total that RoBERTa finds difficult to disambiguate. The dataset contains the following information: it inherits word, lemma, word sense, gloss, and context passage information from WSD datasets. Metaphor labels and overlap scores are added. This dataset covers 82 words and 110 metaphorical word senses. According to the statistics shown in Table \ref{tab:dataset}, we find hard metaphors are quite sparsely distributed in real-life language. Our hard metaphor identification pipeline finds about 110 hard metaphorical senses among 910 metaphorical senses. This is consistent with our expectation in \S\ref{mip} and \S\ref{vuametaphorresult} that hard metaphors are rare among all MIP metaphors. Table \ref{tab:examples} shows several hard metaphor examples with corresponding word sense gloss and overlap ratio $\varphi$. To enable metaphor identification methods to be tested on the hard metaphor dataset, similar to the MIP metaphor corpora, we add literal annotations as negative samples to evaluate the precision of metaphor identification systems. To challenge metaphor identification methods, we choose literal examples that are closest to the metaphor example in the contrastive learning vector space as its negative samples. We maintain the ratio of metaphoric and literal cases to $1:1$ in our dataset. By comparing the cases with different overlap $\varphi$, we  find that $\varphi$ does correlate with our intuition about hard examples (further analysis of overlap score is in \S\ref{downstream result, varphi}).

\begin{table*}[ht]
    \centering
    \resizebox{\textwidth}{!}{
    \begin{tabular}{lp{15.3cm}cc}
    \toprule
        Task & Question & Answer & Pred \\ \midrule
        QA & \textbf{Passage:} A single - volume history has recently been courageously and skilfully attempted by Hugh Honour and John Fleming , which inevitably suffers from the problem of compression . Other civilisations are \textit{\underline{[treated]}} in separate studies . One parallel to the scope of Janson 's book on Western art , \textbf{Question:} Are other civilisations discussed in separate studies ? & T & F \\ \hline
        NLI & \textbf{Sentence 1: } The privatised city centres have taken up too much space.& T & T\\
        &\textbf{Sentence 2: } The privatised city centres have \textit{\underline{[eaten]}} up too much space.  \\ \hline
        MI & \textbf{Sent:} Abraham Lincoln [emerged] as an incarnation of the national Constitution. & M & M \\ 
        & \textbf{Sent:} Moreover, her central figures are so busily fulfilling their multitudinous assignments that none \textit{\underline{[emerges]}} as an arresting individual in his own right or as a provocative symbol of mankind's ills & M & L \\ \hline
        MT & \multicolumn{3}{p{17.5cm}}{ \textbf{Sent: } The world wants to know if Britain can adjust to the facts of life or will allow old fears , old habits , old prejudices , old prides to weigh down its vitality and \textit{\underline{[eat]}} up its resources . \textbf{Reference: } \zh{...并耗尽它的资源...} (..exhaust its resources). \textbf{Prediction: } eat up its resources -> \zh{...吃掉他的资源} (literally eat its resources)} \\ 
    \bottomrule
    \end{tabular}}
    \caption{Examples of downstream tasks for hard metaphors. Answer represents reference answer and Pred indicates model prediction. M means metaphorical and L means literal in MI (metaphor identification) task. MT is English to Chinese machine translation.}
    \label{tab:downstream_example}
\end{table*}

\begin{figure*}[ht]
    \centering
    \includegraphics[width=\textwidth]{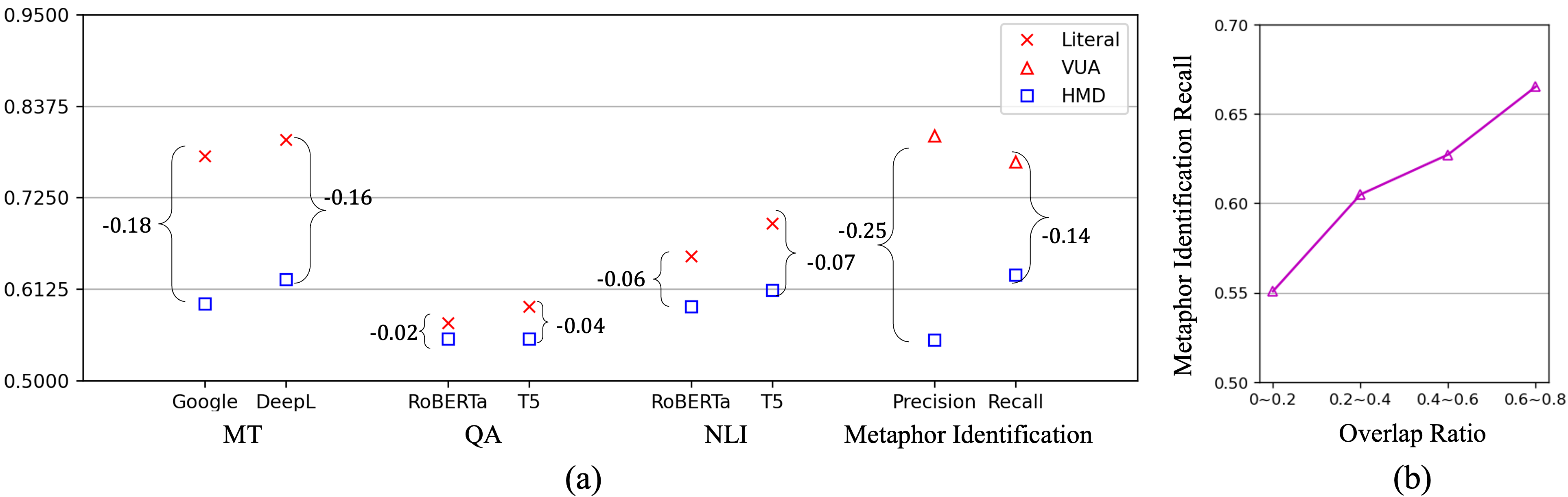}
    \caption{(a) Performance gap between hard metaphors and literal counterparts. (b) Metaphor Identification (MI) Recall score of MelBERT on hard metaphors with different overlap ratio. }
    \label{fig:gap}
\end{figure*}

\section{Analysis of Hard Metaphors}

Here we  analyse the downstream task performance for   the hard metaphors identified for RoBERTa. We also investigate %
how difficult it is to identify hard metaphors.

\subsection{Creating Downstream Task Sets}
\label{downstream_annotation}

To conduct downstream task evaluation, we manually make up  question-answer pairs for hard metaphor examples.
First, we construct the QA test set for the identified hard metaphor dataset that consists of 200 QA examples. Our annotation follows the annotation protocol of BoolQ, a binary passage comprehension questions answer task, that models answer true or false given passages and questions. To ensure the QA test is focusing on the understanding of the metaphors, the annotators were asked to craft questions targeting specifically  the critical metaphor word. Second, we annotate a NLI dataset that consists of 200 sentence pairs constructed with hard metaphors. The annotation of our NLI test set is based on MRPC, a popular NLI task. The labels of our NLI test set are evenly distributed on neutral/contradict/entail. Each pair has an S1 which is an identified hard metaphor as premise, and S2 which is a manually created literal paraphrase as hypothesis. The paraphrase here is constructed based on the label: if the annotator is asked to create a entail/contradict instance, the annotator picks a synonym/antonym of the metaphor word retrieved from WordNet. If synonyms or antonyms are missing or a neutral instance is required, the annotator will choose the word on their own. Annotators followed BoolQ guidelines \cite{clark2019-boolq}, adjusted to focus on the target metaphor in the passage. For the task of English to Chinese machine translation, we add a reference (gold) translation, created manually, but as we conduct manual accuracy evaluation on the machine translation task, the gold translations are not used in our experiments. 
The  annotation task was assigned to college students and professors, where each instance receives at least three annotations. General annotation guidelines and a small amount of reference annotation samples were presented to all annotators. 
We also involved the metaphor identification task in our analysis. Unlike the above three task, this task requires no further human annotation. Examples from the above four tasks can be found in Table~\ref{tab:downstream_example}

\subsection{Baseline Models and Metrics}

\noindent\textbf{Baselines. }
For machine translation (English to Chinese)  %
we used DeepL Translator\footnote{\url{https://www.deepl.com/translator}} and Google Translate\footnote{\url{https://translate.google.com}} as the baseline systems. We use RoBERTa and T5 \cite{raffel2020-t5} in the tests of the two NLU tasks NLI and QA. Note that before the evaluation, both models are fine-tuned on the MRPC \cite{dolan2005-mrpc} and BoolQ \cite{clark2019-boolq} datasets. We choose the SOTA metaphor identifier MelBERT \cite{choi2021melbert} for Metaphor Identification. %

\noindent\textbf{Metrics. }
For translation performance human annotators are required to give a binary judgement, solely for the translation of the metaphor part of the sentence. The annotators and guidelines are the same as in \S\ref{downstream_annotation}.
We use accuracy as well in NLI and QA tasks and report both precision and recall scores for metaphor identification.

\subsection{Results}
\label{downstream result, varphi}

Fig.~\ref{fig:gap} shows noticeable performance drops among all downstream tasks resulting from the hard metaphors. %
The most significant drop is $0.25$ in the precision of metaphor identification. This means MelBERT has many false positives, but it is largely because  MelBERT was trained to identify MIP metaphors, which tends to label all cases as metaphoric except these expressing very basic meaning (see `basic meaning' explanation in Sec.~\ref{bm}). Recall is a fairer measure here because our HMD is a subset of MIP metaphors (and hence recall is used in Fig.~\ref{fig:gap} (b)). The 0.14 decrease in recall score on metaphor identification shows that hard metaphors are harder to detect than MIP metaphors. For English to Chinese translation there is a decrease of $0.18$ when moving from literal samples to metaphor samples, for translation using Google translate.  DeepL performs slightly better than Google, with a $0.16$ drop.  For the NLI task using RoBERTa and T5 fine-tuned with the BoolQ dataset, hard metaphor causes a 6-7\% drop. The smallest gap, 2-3\%, between hard metaphor and literal counterparts, is on the QA task. It is mainly because the base performance of QA for literal expressions is quite low, so  even though the performance for hard metaphors is worse, the gap is not that noticeable.  A significance test (two-tailed t-test) yielded $p< 0.01$ for QA and NLI tasks, and $p < 0.001$ for the machine translation (English to Chinese) and metaphor identification tasks.

Table \ref{tab:downstream_example}  shows model prediction examples. In some cases, baseline models give correct answers, but baselines fail to solve many of them due to the hard metaphorical meaning. The example  machine translation task in the table is representative. Google translate misunderstood the metaphor as its literal meaning, producing a text in the target language which does not fit  the context.

To investigate the relationship between the overlap ratio $\varphi$ and the difficulty of   metaphor identification, we calculate the recall score of the metaphor identification task for  metaphors with different overlap ratios. The results in Fig.~\ref{fig:gap} (b) show a clear positive correlation between  recall and overlap score. We see that as the overlap ratio approaches zero, the recall gets  close to 0.5 (the performance of random guessing). This  validates our hypothesis: \textit{metaphors having poor overlap in PLM embedding space are hard to identify}. %

\begin{figure}
    \centering
    \includegraphics[width=\columnwidth]{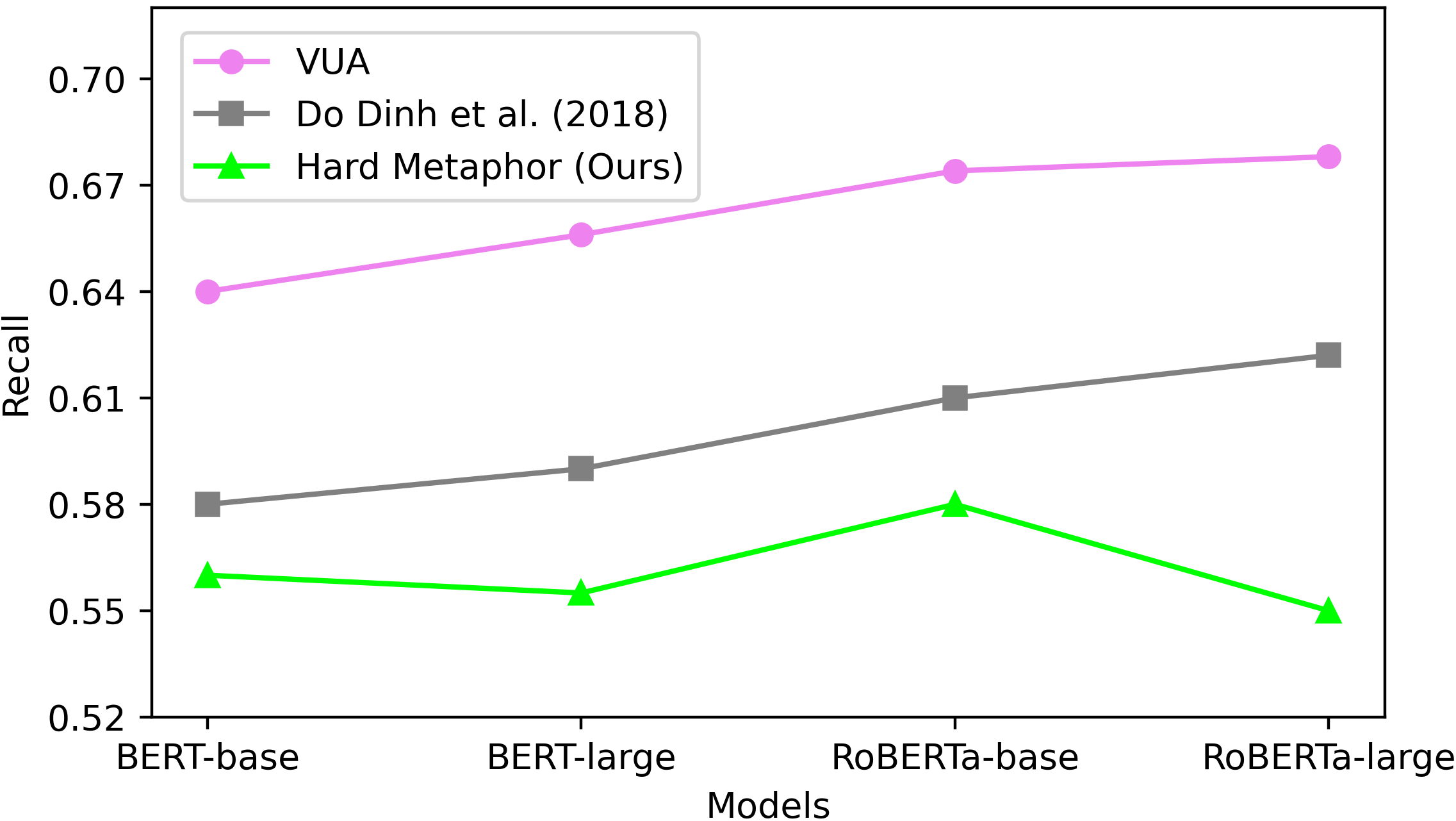}
    \caption{
    Comparison of recall in metaphor identification for three different datasets. The overlap threshold in our hard metaphor construction is set to 0.5. The trend shows that VUA and \citet{do-dinh-etal-2018-weeding} are gradually  becoming solved as models improve. 
    }
    \label{fig:dodinh}
\end{figure}

\subsection{Context of the Metaphor}
Manual inspection of our hard metaphors reveals some word senses that would be considered novel, but also many that seem mundane. For many metaphorical word senses
we find a huge variance in overlap score, sometimes spanning the whole range from zero to 0.8 (the threshold for hard metaphor). These are words with the same metaphorical sense, the variance being due to the context in which they occur. We did manual analysis of the contexts of samples with very different overlap scores, and we found it very difficult to see why they have such different overlap scores for the LLM. For example, for the sense of `arm' being `any projection that is thought to resemble an arm', we have a sample including `furniture was piled in high \ldots  a leg discernible here, an upholstered [arm] protruding', that scores zero overlap. Samples at the high end of overlap scores mention machine arms in contexts that do not seem easier to a human. 
It is known that LLMs can be extremely context sensitive, often in ways that seem surprising to humans, for example \citet{bubeck2023sparks}  noted that ``changes in the wording of the question can alter the knowledge that the model displays.'' Humans are less sensitive to superficial variations in context, and seem to be better able to grasp the underlying meaning. 
Based on samples we have analysed we feel that a human annotator cannot identify  metaphors that would be hard for an  LLM. 

\subsection{Comparison with Other Challenging Metaphor Datasets}

\citet{do-dinh-etal-2018-weeding} developed a dataset that is a subset of VUA without auxiliary verbs and prepositions, and providing novelty annotations to allow researchers to focus  on novel metaphors (as opposed to conventional/dead metaphors). We tested the  difficulty of metaphor identification on their novel metaphors, versus VUA, and our hard metaphors, in Fig.~\ref{fig:dodinh}. Note that all models are fine-tuned on the VUA training set with same hyper-parameters before they are tested. \citeauthor{do-dinh-etal-2018-weeding}'s mertaphors are indeed harder to identify than VUA (lower recall, see reason for using recall in Sec.~\ref{downstream result, varphi}), however the trend of increasing performance with more sophisticated models shows that such a dataset will become outdated with time. It is not clear how it should be updated as more sophisticated models come, because its method is based on human opinion of what is a novel metaphor.
In contrast, our method  can find the weakness of a statistical model, which means our hard metaphors are tailored for each model, providing a dataset that will challenge it.

\citet{neidlein-etal-2020-analysis} also showed that hardness for models to identify metaphors is correlated with \citeauthor{do-dinh-etal-2018-weeding}'s novelty measure. \citeauthor{neidlein-etal-2020-analysis} find this result counter-intuitive, because humans would easily recognise a word being used in quite an unusual way. However simple statistics may explain the model behaviour: models rely on seeing plentiful training samples.  \citeauthor{do-dinh-etal-2018-weeding} did find a moderate inverse correlation between frequency of words and novelty. 

The above works  place much emphasis on the metaphor word itself, 
with \citeauthor{neidlein-etal-2020-analysis} additionally analysing inflectional variations and  synonyms of metaphor words.
However they neglect the contribution of context. We find in our hard metaphor dataset that the context is  critical to hardness for computers. 
Even if a metaphor word sense is frequent, if it appears in very varied contexts, it will upset a model's efforts to learn the contexts which imply that sense. 
Relying on human opinion of the novelty of a metaphor misses out on recording the novelty and complexity of the context in which it is used. 

FLUTE \cite{chakrabarty2022flute} is another dataset on figurative language, including metaphors, and focusing on NLI. It is a high quality dataset, including text explanations. IMPLI \cite{stowe-etal-2022-impli} is a similar dataset, but without explanations. Both of these do not contain word sense annotations, and unlike \citet{do-dinh-etal-2018-weeding} they are not concerned with the novelty of metaphors, nor are they concerned with hardness of metaphors, as is our main focus. Therefore they contain many common metaphors such as `drop prices' \cite{stowe-etal-2022-impli}, which would not trouble typical LLMs, because they have seen common metaphors like this frequently in training. These datasets could be useful   as a base dataset for our method. If word sense annotations can be added, then we could select the hard metaphors from their larger set. It can be expected that some hard metaphors are also in these sets.

\subsection{Discussion}

\textbf{Hard metaphors as a benchmark for metaphor processing. } 
Our hard metaphor dataset mainly focuses on metaphors that SOTA NLP methods fail to tackle. Considering the statistical nature of existing NLP methods, this is equivalent to finding a long-tail subset of metaphors from a broad MIP metaphor corpus. It is the rareness or long-tail nature of hard metaphors that makes them a challenge for NLP models. It is not necessarily the metaphoric word sense that is rare, but its use in a specific context. The long-tail nature of hard metaphors makes it hard to address by simply adding more training data, as the marginal utility of adding training samples diminishes. Therefore, we recommend our hard metaphor dataset to be used as a benchmark rather than a training set for metaphor processing methods. We also recommend symbolic approaches to be tested on our hard metaphor dataset as they are promising methods especially for long-tail scenarios.

\noindent \textbf{Hard metaphor dataset has more possibilities than VUA. } Although VUA is the largest and most popular metaphor corpus, its usage is quite limited. It provides only binary metaphoric labels for each word, which makes it a pure metaphor identification dataset. In contrast, HMD  contains the  metaphorical meaning of each hard metaphor in the form of WordNet sense ID and gloss, which supports further studies on metaphor interpretation. In addition, the overlap ratio as an indicator for difficulty/rareness has the potential to facilitate  exploration of creative metaphors. The metaphor and literal test samples we have made for downstream tasks also enable metaphor processing methods to be tested in real NLP tasks.

\noindent\textbf{Why metaphors are special.} Our contrastive learning and overlap ratio  approach is a generic method. It can be used to find not only hard metaphors but also any word sense in a context, that is hard to identify. %
Here we have focused only on metaphors because we believe metaphors are special:
Humans are good at understanding metaphors, even novel cases that they have never seen before, because they can create a mapping among the concepts involved. But machines struggle to tackle rare creative metaphors due to the machine learning approach relying on statistics. Based on this observation, we hope that our hard metaphor dataset can encourage the community to explore how humans tackle rare or long-tail metaphors, which might involve commonsense reasoning and symbolic inference.

\section{Conclusion}

Although existing metaphor processing methods, especially metaphor identification models, have made significant progress, most of them focus on tackling MIP metaphors which has little impact on real NLP tasks. 
To address this we proposed an automatic approach to detect hard metaphors that current SOTA approaches fail on, and introduced the hard metaphor dataset tailored for RoBERTa. Moreover, we present the first metaphor corpus with downstream question-answer pairs which enables metaphor processing methods to be tested for their potential on real NLP applications. We found that what makes a metaphor hard for an LLM is the context in which it is used; the novelty of the metaphor, as judged by a human, is not a good indicator of its difficulty for an LLM. 

\clearpage

\bibliography{tacl2021}
\bibliographystyle{acl_natbib}

\end{document}